%% file: example_paper.tex
\documentclass[nohyperref]{article}

\usepackage{microtype}
\usepackage{pifont}
\usepackage{graphicx}
\usepackage{subfigure}
\usepackage{booktabs} 
\usepackage{hyperref}
\usepackage{amsmath}
\usepackage{amssymb}
\usepackage{amsthm}
\usepackage{mathtools}
\usepackage[accepted]{icml2022}

\usepackage[capitalize,noabbrev]{cleveref}
\usepackage[textsize=tiny]{todonotes}

\theoremstyle{plain}

\theoremstyle{definition}

\theoremstyle{remark}

\icmltitlerunning{AnyMorph: Learning Transferable Polices By Inferring Agent Morphology}

\begin{document}

\twocolumn[
\icmltitle{AnyMorph: Learning Transferable Polices By Inferring Agent Morphology}

\icmlsetsymbol{equal}{*}

\begin{icmlauthorlist}
\icmlauthor{Brandon Trabucco}{cmu}
\icmlauthor{Mariano Phielipp}{equal,intel}
\icmlauthor{Glen Berseth}{equal,mila}
\end{icmlauthorlist}

\icmlaffiliation{cmu}{Machine Learning Department, Carnegie Mellon University, work done while at Intel AI}
\icmlaffiliation{intel}{Intel AI}
\icmlaffiliation{mila}{Mila}

\icmlcorrespondingauthor{Brandon Trabucco}{btrabucco@cmu.edu}

\icmlkeywords{Deep Learning, Transformers, Reinforcement Learning}

\vskip 0.3in
]

\printAffiliationsAndNotice{\icmlEqualContribution}

\begin{abstract}
\input{sections/abstract}
\end{abstract}

\section{Introduction}
\label{sec:intro}
\input{sections/intro2}

\section{Related Works}
\label{sec:related_works}
\input{sections/related_works}

\section{How Do You Represent Morphology?}
\label{sec:morphology}
\input{sections/motivations}

\section{Morphology As Sequence Modelling}
\label{sec:method}
\input{sections/methodology}

\section{Training Performance}
\label{sec:experiments_one}
\input{sections/experiments_one}

\section{Zero-Shot Generalization}
\label{sec:experiments_two}
\input{sections/experiments_two}

\section{Visualizing Robustness}
\label{sec:experiments_three}
\input{sections/experiments_three}

\section{Conclusion}
\label{sec:conclusion}
\input{sections/conclusion}

\section{Acknowledgements} This work was supported by Intel AI and CIFAR. We thank Benjamin Eysenbach for providing feedback on an early draft of the paper and for fruitful discussions about prior work. We thank Sergey Levine, Ruslan Salakhutdinov, and Deepak Pathak for providing feedback on the paper and for helpful discussions about the approach and how to evaluate it. We thank the authors \citet{DBLP:conf/icml/HuangMP20} for providing an open source benchmark and an implementation of SMP. We thank the authors \citet{DBLP:conf/iclr/KurinIRBW21} for providing an implementation of Amorpheus, which we used as a starting point. We thank Wenlong Huang for help debugging the SMP baseline. We thank Vitaly Kurin for pointing us to new papers in agent-agnostic RL that helped our literature review. We thank the anonymous reviewers at ICML for providing detailed feedback on a draft of the paper.

\bibliography{example_paper}
\bibliographystyle{icml2022}

\newpage
\appendix
\onecolumn

\input{sections/appendix}

\end{document}

%% file: sections/abstract.tex
The prototypical approach to reinforcement learning involves training policies tailored to a particular agent from scratch for every new morphology.
Recent work aims to eliminate the re-training of policies by investigating whether a morphology-agnostic policy, trained on a diverse set of agents with similar task objectives, can be transferred to new agents with unseen morphologies without re-training. This is a challenging problem that required previous approaches to use hand-designed descriptions of the new agent's morphology. Instead of hand-designing this description, we propose a data-driven method that learns a representation of morphology directly from the reinforcement learning objective.
Ours is the first reinforcement learning algorithm that can train a policy to generalize to
new agent morphologies without requiring a description of the agent's morphology in advance. We evaluate our approach on the standard benchmark for agent-agnostic control, and improve over the current state of the art in zero-shot generalization 
to new agents. Importantly, our method attains good performance \textit{without} an explicit description of morphology.

%% file: sections/intro2.tex
\begin{figure}
    \centering
    \includegraphics[width=\linewidth]{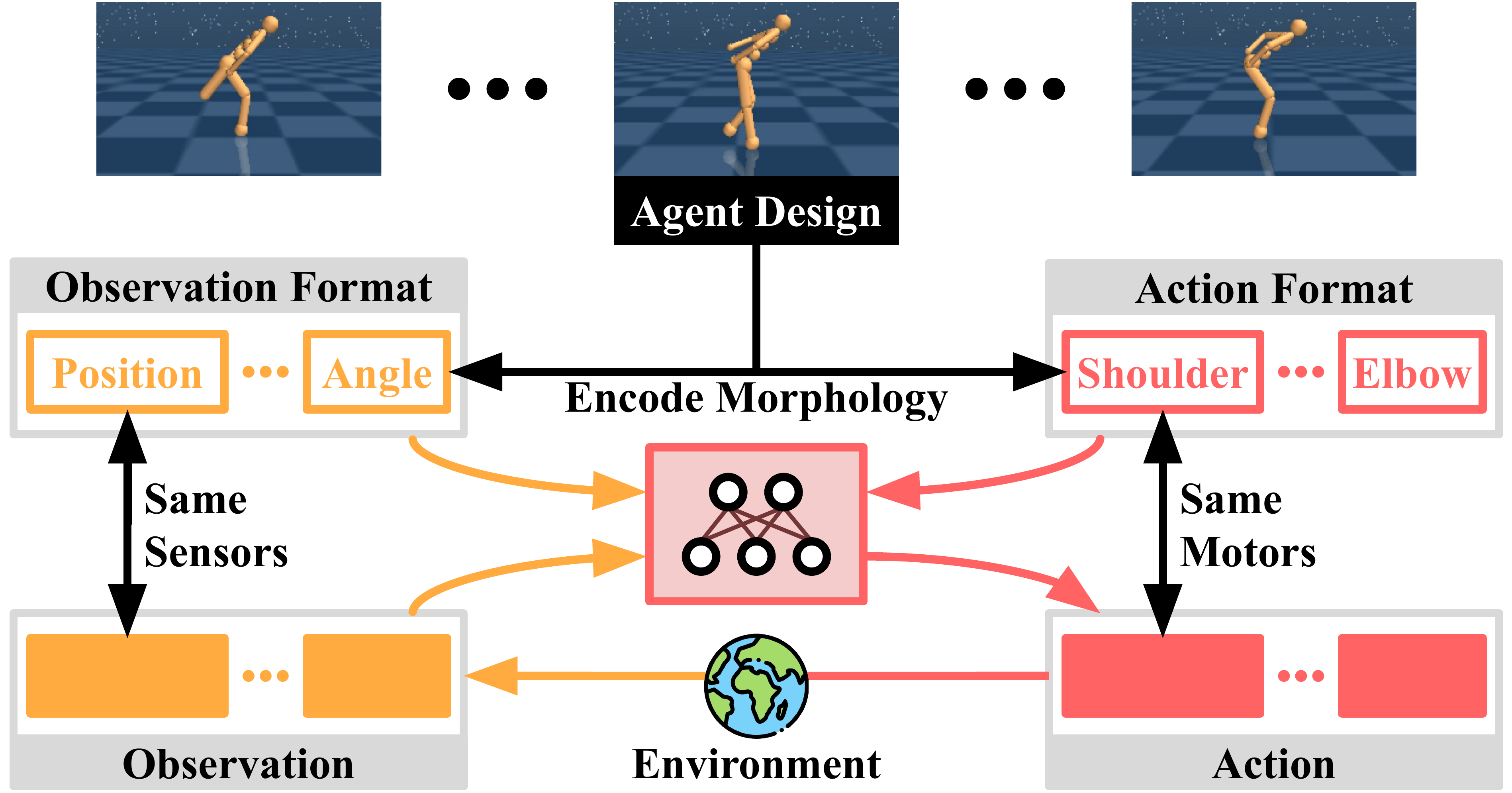}
    \vspace{-0.5cm}
    \caption{Overview of the architecture of our policy for an example reinforcement learning task with three agent morphologies. The agent's morphology is represented by a sequence of tokens, and is processed by a sequence-to-sequence Transformer policy.}
    \vspace{-0.6cm}
    \label{fig:policy_architecture}
\end{figure}

Agent-agnostic reinforcement learning (RL) is an emerging research challenge that involves training policies that are transferable to new agents with different morphology. Rather than training policies from scratch for every new agent, a pretrained agent-agnostic policy can provide an effective solution with potentially no additional training. The core philosophy motivating agent-agnostic RL is a notion that large multi-task models exhibit strong generalization behavior. This result is continually observed in domains outside robotics, where models like BERT \cite{DBLP:conf/naacl/DevlinCLT19}, GPT-3 \cite{DBLP:conf/nips/BrownMRSKDNSSAA20}, and CLIP \cite{DBLP:conf/icml/RadfordKHRGASAM21} exemplify how large multi-task models zero-shot generalize in out-of-domain applications like CLIP in robotics \cite{DBLP:journals/corr/abs-2111-09888}.
Agent-agnostic RL has the potential to create policies with comparable generalization to these foundation models \cite{DBLP:journals/corr/abs-2108-07258}. 
However, agent-agnostic RL is hard because it requires an effective representation of the agent's morphology  \cite{DBLP:conf/icml/HuangMP20}, which remains a challenging research problem. The morphology representation is crucial in agent-agnostic reinforcement learning because it must contain all the information about the agent's physics needed for control. In this work, we consider how to learn this representation instead of manually designing it, as is standard in current research.

 
Existing works have approached agent-agnostic RL by assuming the agent obeys strict design criteria. First, the agent has limbs. Second, each limb has similar proprioceptive sensors \cite{DBLP:conf/iclr/WangLBF18, DBLP:conf/icml/HuangMP20, DBLP:conf/iclr/KurinIRBW21}. These assumptions restrict agent-agnostic RL to rigid-body agents in a MuJoCo-like \cite{DBLP:conf/iros/TodorovET12} simulator that exposes congruent limb-observations. Deviating from prior work, we consider a setting that is more flexible than prior work. Our method does not require the agent to have limbs, nor does it require observations to be annotated according to which limb they describe.
Instead, our method automatically infers how the agent's sensors and motors are connected from the RL objective, illustrated in Figure~\ref{fig:policy_architecture}. Our approach performs 16\% better (see Section~\ref{sec:experiments_two}) than existing methods in zero-shot generalization to new morphologies, and requires \textit{less} information about the morphology of the agent being controlled than prior art.

In this paper, we make the following contributions. First, we frame learning morphology as a sequence modelling problem, and represent an agent's morphology as a sequence of tokens, with corresponding learnt embeddings. In this fashion, an agent's morphology is differentiable, and is optimized from the reinforcement learning objective. Second, we propose an agent-agnostic neural network architecture that generalizes effectively and does not assume the agent has limbs, nor has per-limb observations. 
Third, we demonstrate that our approach generalizes up to $32\%$ better than existing work on large tasks in a standard benchmark,
and displays an emergent robustness to broken sensors.

\begin{table*}[htbp]
    \centering
    \vspace{-0.2cm}
    \caption{Comparison of the structural assumptions that prior works make about the decision-making agent. Our method is more flexible than prior work, requiring neither graph structure, nor explicitly aligning the agent's sensors and actuators to its limbs in order to generalize effectively. Our approach requires less information about the agent's morphology than prior work, and performs $16\%$ better in zero-shot generalization to new morphologies than existing methods with stronger assumptions, which is discussed in Section~\ref{sec:experiments_two}.}
    \vspace{0.5cm}
    \begin{tabular}{l|c|c}
        \toprule
        \textbf{Method \textbackslash \;Assumptions} & \textbf{Graph Structure} & \textbf{Alignment}\\
        \midrule
        Graph Networks \cite{DBLP:conf/icml/Sanchez-Gonzalez18} & \ding{55} & \ding{55} \\
        NerveNet \cite{DBLP:conf/iclr/WangLBF18} & \ding{55} & \ding{55} \\
        Shared Modular Policies (SMP) \cite{DBLP:conf/icml/HuangMP20} & \ding{55} & \ding{55} \\
        Amorpheus \cite{DBLP:conf/iclr/KurinIRBW21} & & \ding{55} \\
        \midrule
        AnyMorph (Ours) & & \\
        \bottomrule
    \end{tabular}
    \label{tab:assumptions}
\end{table*}

%% file: sections/related_works.tex
Generalization in reinforcement learning has a rich history, with early works demonstrating generalization to new tasks in robotics \citet{Sutton2011Horde}, and in video games \cite{Schaul2015uvfa, DBLP:journals/corr/ParisottoBS15}. Tasks were often defined via goal states \citet{Sutton2011Horde}, with a reinforcement learning objective defined as minimizing the distance to the goal throughout an episode \cite{DBLP:conf/nips/AndrychowiczCRS17, DBLP:conf/nips/NasirianyPLL19}. These goal-based methods can be improved combining them with language \cite{DBLP:conf/nips/JiangGMF19}, and learnt latent spaces \cite{DBLP:conf/iclr/EysenbachGIL19, DBLP:conf/icml/RakellyZFLQ19}. As the field matures, researchers are beginning to investigate larger-scale multi-task reinforcement learning (MTRL)~\cite{MTRLsurvey2020}, including generalization across all of Atari \cite{DBLP:conf/iclr/HafnerLB020}, and generalization to new agents with different dynamics or morphology. The latter is an emerging topic called agent-agnostic reinforcement learning \cite{DBLP:conf/icra/DevinGDAL17, DBLP:conf/icml/HejnaPA20, DBLP:conf/icml/HuangMP20}. 
This setting is challenging because different agents typically have incompatible (different cardinality) observations and actions, which precludes conventional deep reinforcement learning approaches. 

Conventional deep reinforcement learning approaches expect fixed-size observations and actions, and devising effective alternatives is an open research challenge. Existing work has shown modularization can both improve multi-task generalization \cite{DBLP:conf/icml/0003KLG21, DBLP:conf/iclr/GoyalLHSLBS21},
and allow processing of different-cardinality observations and actions \cite{DBLP:conf/icml/HuangMP20}. Many such approaches utilize Graph Neural Networks \cite{gnns2005, gnnsForWebRanking2005}, conditioning the policy on a graph representation of the agent's morphology. Graph-based approaches have demonstrated the ability to generalize to novel agent morphology \cite{DBLP:conf/icml/HuangMP20}, and learn complex gaits \cite{DBLP:conf/iclr/WangLBF18, DBLP:conf/icml/Sanchez-Gonzalez18}. However, recent work has shown that generalize can be further improved using Transformers \cite{DBLP:conf/iclr/KurinIRBW21}, due to their success in modelling dynamic structure via the attention mechanism \cite{tenney2018context, DBLP:conf/blackboxnlp/VigB19}. Agent-agnostic reinforcement learning methods, including those with Transformers, currently rely on a manually-designed representation of morphology (see Section~\ref{sec:morphology}). Our approach differs from prior work by instead learning a representation of morphology with reinforcement learning that produces better generalization than a manually-designed representation.


%% file: sections/motivations.tex

The goal of this section is to outline what information is needed to define a \textit{morphology}. This terminology is used, for example, to denote an interpretation of the physical rigid-body of a MuJoCo-like \cite{DBLP:conf/iros/TodorovET12}
agent as a graph, where nodes correspond to rigid limbs, and edges indicate two limbs are connected by a joint. 
This terminology is based on classical work on kinematic chains~\citep{denavit1955kinematic}.
In this section, we will explore a standard definition of morphology for decision-making agents used in existing work, and will discuss its benefits and limitations.
Towards this goal, we first introduce the reader to notation that will be referred to throughout our paper.

\paragraph{Preliminaries.} We take inspiration from and modify the notation presented by \citet[p. 3]{DBLP:conf/icml/HuangMP20}. Consider a set of $N$ decision-making agents with different observation spaces and actions spaces---to avoid a recursive definition of \textit{morphology}, we intentionally avoid using it here. 
For each agent $n \in \{ 1 \hdots N \}$, there is a corresponding Markov Decision Process $\mathcal{M}_{n}$ with observation space $S_n$, action space $A_n$, transition dynamics $p_n$, and reward function $r_n$. We will assume for notational convenience that each observation $\mathbf{s} \in S_n$ can be expressed as a vector with $N_{S}(n)$ elements, and that each action $\mathbf{a} \in A_n$ can be expressed as a vector with $N_{A}(n)$ elements.
The goal of agent-agnostic reinforcement learning is to find a policy that maximizes the expected discounted return for the expected agent.
\begin{equation}
\max_{\pi} \; \mathbb{E} \left[ \;\sum_{t = 0}^{\infty} \gamma^{t} r_{n}(\mathbf{s}_t, \mathbf{a}_t, \mathbf{s}_{t+1}) \right]
\end{equation}
The optimal policy $\pi^{*} (\mathbf{a}_t | \mathbf{s}_t, n)$ for this objective depends on the morphology of the current agent $n$. 
How the morphology of the current agent is represented from the perspective of the policy is an important question in our investigation. Let's first identify how prior work answers this question.

\begin{figure}[h]
    \centering
    \includegraphics[width=\linewidth]{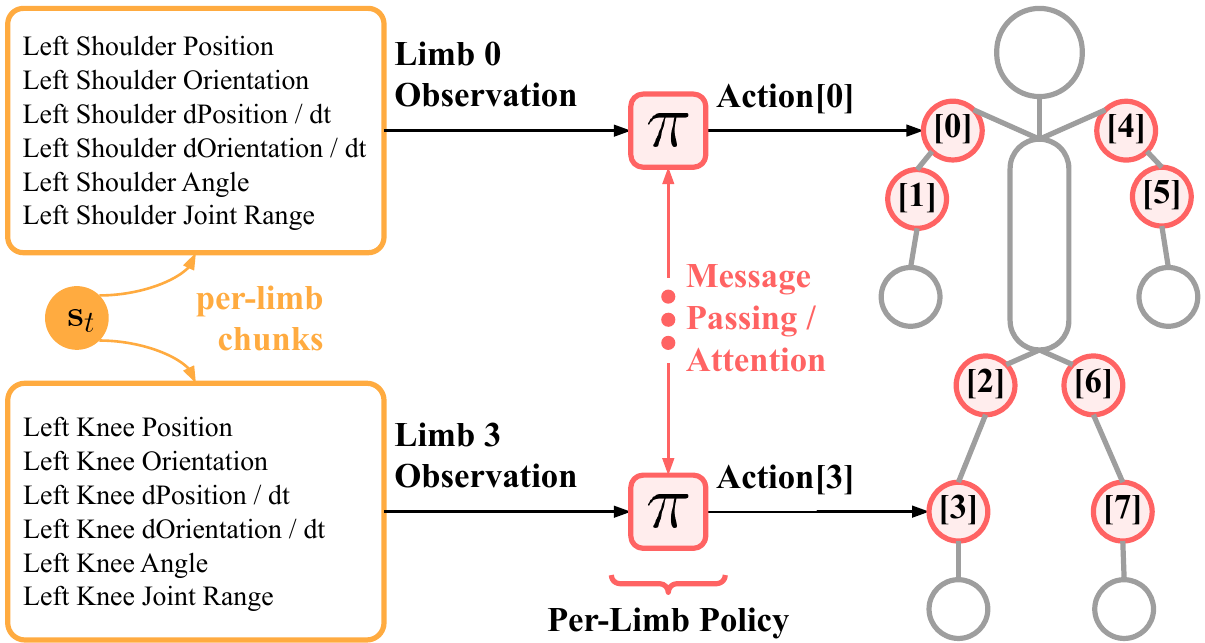}
    \vspace{-0.5cm}
    \caption{Illustration of how existing methods partition the action space and observation space of the \texttt{humanoid\_2d\_9\_full} morphology. These works assume limbs are uniformly designed, and require additional annotations about how sensors and motors are connected to limbs ($P_{A}$ and $P_{S}$) before learning can happen.}
    \label{fig:example_morphology}
\end{figure}

\paragraph{Using Graph-based Morphologies.} The \textit{morphology} of a decision-making agent is often represented as a graph. Consider the agent-conditioned undirected graph $\mathcal{G}_n = (V_n, E_n)$ whose vertices $V_n$ each represent a limb the agent has, and whose set of edges $E_n$ contains all pairs of limbs that are connected by a joint. For the typical MuJoCo-like \cite{DBLP:conf/iros/TodorovET12} 
agent from prior work, the topology of this graph closely resembles the topology of the agent's body.
\begin{gather}
V_n = \left\{ 1 \hdots N_{L}(n) \right\} \\
E_n = \left\{ (i,j) : \text{$i$ and $j$ are connected limbs} \right\}
\end{gather}
Graph-based morphologies are intuitive because they mirrors how the agent is physically connected, but $\mathcal{G}_n$ is not a complete description of morphology. The number of limbs $N_{L}(n)$ is generally independent of the dimensionality of the action space $N_{A}(n)$ and observation space $N_{S}(n)$. The agent must also know which individual sensors and motors are connected to each limb. This knowledge is typically specified via an $N_{L}(n)$-way partition $P_{A}$ of the action space  and an $N_{L}(n)$-way partition $P_{S}$ of the observation space. Figure~\ref{fig:example_morphology} shows how sensors are motors are partitioned in prior work for an example humanoid morphology. 

\begin{figure*}
    \centering
    \includegraphics[width=\linewidth]{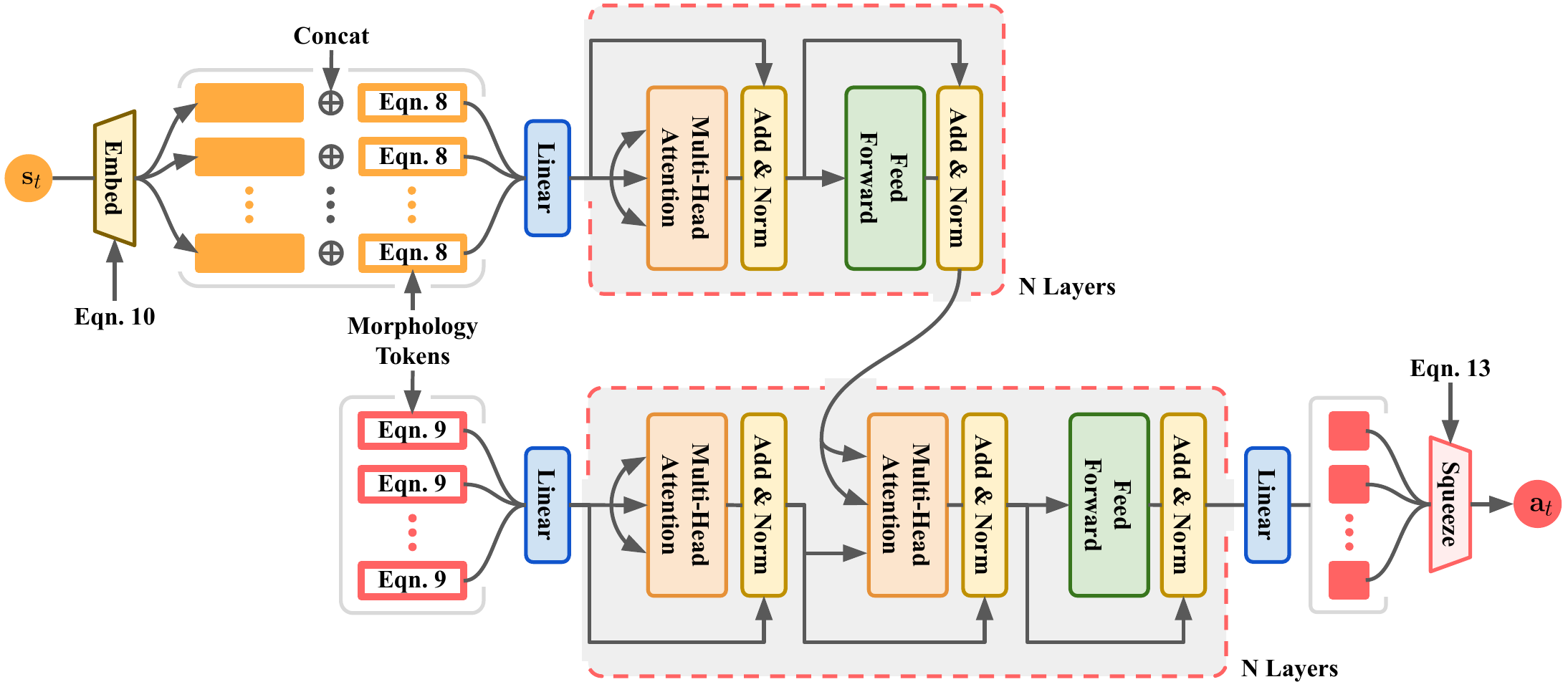}
    \vspace{-0.5cm}
    \caption{Visualization of the architecture of our policy. Our policy is a sequence-to-sequence deep neural network, consisting of a Transformer \cite{DBLP:conf/nips/VaswaniSPUJGKP17} encoder that processes the current state $\mathbf{s}_t$ and a sequence of embeddings of observation tokens (see Section~\ref{sec:method}), with a non-masked Transformer decoder that processes a sequence of embeddings of action tokens, given the encoder hidden state. Our policy is invariant to the dimensionality of observations and actions, and does not assume the agent has limbs.}
    \label{fig:main_diagram}
\end{figure*}



\paragraph{Consequences Of Graph-based Morphologies.} One appealing quality of these graph-based morphologies are that they have an intuitive physical interpretation. Given only the morphology, one can invert the definition to recover the agent's physical design, including the arrangement of its sensors. This appealing quality comes at the expense of strict design criteria, however.
In this section, we will discuss several consequences of the graph definition of morphology. 

\begin{itemize}
    \item First, this definition assumes the agent's physical design can be expressed as a graph, which restricts methods to MuJoCo-like \cite{DBLP:conf/iros/TodorovET12} agents with rigid limbs, which is not true, for example, for a car with flexible rubber tires. We refer to this as the \textbf{graph structure assumption} in Table~\ref{tab:assumptions}.
    The community is interested in agent-agnostic RL for real-world robotic systems, and a graph inductive bias may be insufficient to represent the complex physique of certain agents.
    \item Another common assumption is that sensors and actuators are categorizable according to a particular limb. 
    Methods with this \textbf{alignment assumption} in Table~\ref{tab:assumptions} only maintain agnosticism to limbs, and not to individual sensors and actuators \cite{DBLP:conf/icml/HuangMP20, DBLP:conf/iclr/KurinIRBW21}. Some interesting reinforcement learning problems violate
    this property by not having per-limb sensors, or even well-defined limbs. Tackling these problems may require more flexible agent-agnostic methods that do not require aligning sensors to limbs. 
\end{itemize}

%% file: sections/methodology.tex

Defining morphology in terms of limbs, which we discussed in Section~\ref{sec:morphology}, often requires making three unrealistic assumptions about the agent (see Table~\ref{tab:assumptions}). In this section, we will present an alternate representation of morphology for general decision-making agents that \textit{outperforms} existing methods while being applicable to a larger space of agent types.
We approach this problem by considering two algorithmic desiderata: for a given agent $n$, (1) our method should be invariant to the dimensionality of the agent's observations and actions, and (2) our method should only be given \textit{minimal} information about the agent's morphology. Our first insight is to interpret the reinforcement learning policy that accepts a state vector and outputs an action vector as sequence-to-sequence mappings, where the source sequence has $N_{S}(n)$ elements, and the target sequence has $N_{A}(n)$ elements. This interpretation bypasses all dependence on limbs by defining morphology in terms of which sensors and actuators comprise the agent. Inspired by the success of word embeddings \cite{DBLP:journals/corr/abs-1301-3781} in natural language processing, and their role in large-scale models like BERT \cite{DBLP:conf/naacl/DevlinCLT19}, we propose to encode sensors and actuators as a sequence of tokens, and represent them with learnt embeddings. 
We learn two embedding that represent observation and action tokens respectively.
\begin{gather}\label{eqn:obs_act_embeddings}
    H_n^{\;\text{obs}} \in \mathbb{R}^{N_{S}(n) \times D}, \; H_n^{\;\text{act}} \in \mathbb{R}^{N_{A}(n) \times D}
\end{gather}
Each token encodes the identity of a single sensor or actuator. Embeddings representing these tokens are learned jointly with our policy directly from the reinforcement learning objective. Our policy is a sequence-to-sequence deep neural network that maps from a source sequence with $N_{S}(n)$ elements to a target sequence with $N_{A}(n)$ elements. We implement our policy as a variant of the Transformer \cite{DBLP:conf/nips/VaswaniSPUJGKP17} due to its success in multiple reinforcement learning settings \cite{DBLP:conf/iclr/KurinIRBW21, janner2021offline, chen2021decision}. In the following section, we describe our architecture (shown in Figure~\ref{fig:main_diagram}), and modifications that make it suitable for agent-agnostic reinforcement learning.

\begin{figure*}[htbp]
    \centering
    \includegraphics[width=\linewidth]{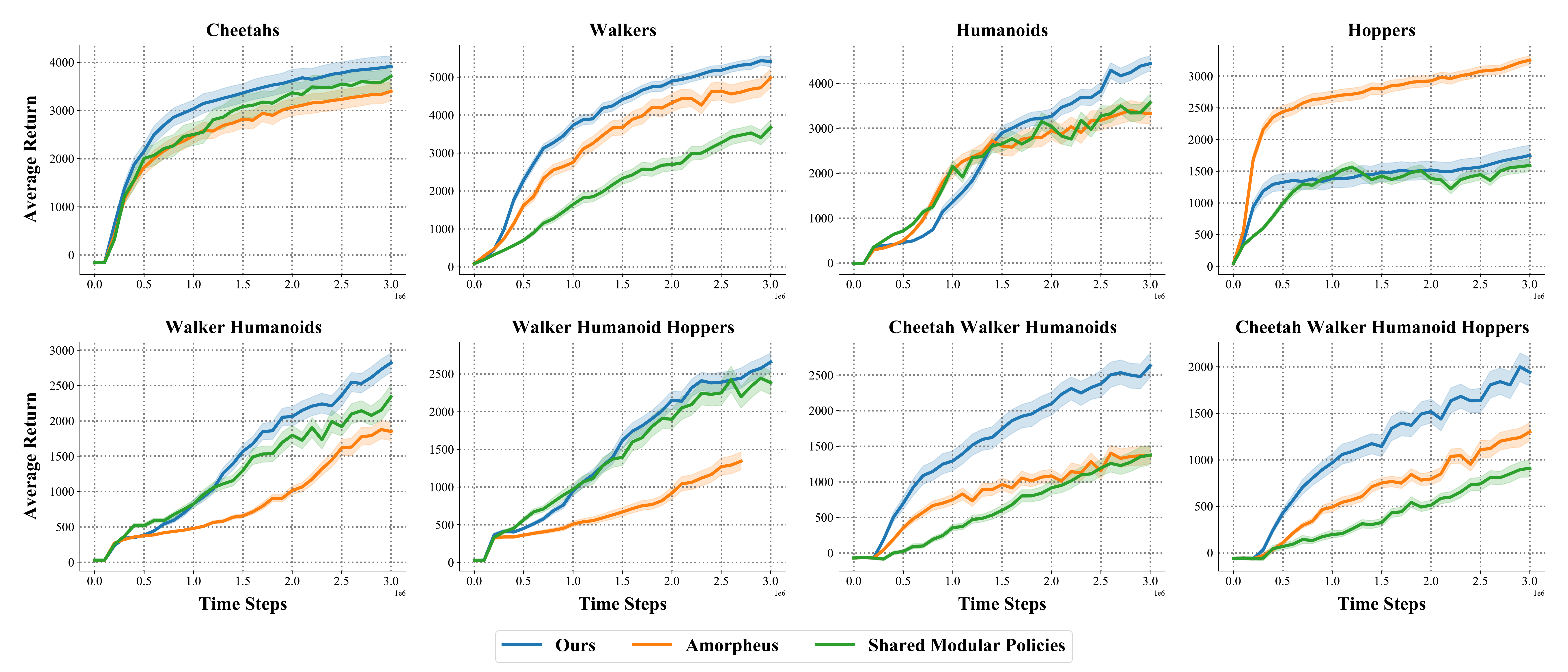}
    \vspace{-0.5cm}
    \caption{Average return of our method versus Amorpheus~\citep{DBLP:conf/iclr/KurinIRBW21} and Shared Modular Policies~\citep{DBLP:conf/icml/HuangMP20} across $4$ random seeds. Performance on the y-axis is the average return with one episode per training morphology per seed. Dark colored lines indicate the average training performance, and a $95\%$ confidence interval is shown with shading around each line. The x-axis indicates the total steps across all training morphologies. Our method frequently reaches and exceeds the performance of baselines, improving by $85\%$ and $53\%$ on Cheetah-Walker-Humanoids and Cheetah-Walker-Humanoid-Hoppers respectively, the tasks with the most morphologies.
    }
    \label{fig:section_one}
\end{figure*}

\paragraph{Morphology Tokens.} Before detailing our policy architecture, we will explore how the necessary observation embeddings $H_n^{\;\text{obs}}$ and action embeddings $H_n^{\;\text{act}}$ can be obtained. In the previous section, we described a corresponding set of observation tokens $I_{n}^{\;\text{obs}}$ and action tokens $I_{n}^{\;\text{act}}$ that are represented by these embeddings. These tokens are consecutive integers that uniquely index the sensors and motors comprising the agent's action space and observation space. Since the action space has dimensionality $N_A(n)$, the action tokens $I_{n}^{\;\text{obs}}$ can take at most $N_A(n)$ possible values. Similarly, the observation tokens $I_{n}^{\;\text{obs}}$ can take on at most $N_S(n)$ possible values. Emphasizing our inspiration from language modelling, the pair $(I_{n}^{\;\text{obs}}, I_{n}^{\;\text{act}})$ functions like word tokens \citep{DBLP:journals/corr/abs-1301-3781}. To honor this inspiration from language modelling, we label this pair \textit{morphology tokens}.
\begin{equation}
    I_{n}^{\;\text{obs}} \in \mathbb{N}^{N_S(n)}, \; I_{n}^{\;\text{act}} \in \mathbb{N}^{N_A(n)}
\end{equation}
Just as word tokens specify an index into an associated set of word embeddings \citep{DBLP:journals/corr/abs-1301-3781}, morphology tokens specify an index into a corresponding pair of morphology embeddings $W_{e}^{\;\text{obs}}$, and $W_{e}^{\;\text{act}}$. These morphology embeddings are weight matrices with $D$ columns. Each weight matrix learns an embedding for each sensor and motor in the observation and action spaces of every morphology, similar to how word embeddings learn an embedding for each word in a vocabulary. See Appendix~\ref{app:tokens} for a discussion of how our morphology tokens are selected in practice.
\begin{gather}
    W_{e}^{\;\text{obs}} \in \mathbb{R}^{ \left( \prod_{i = 1}^N N_S(i) \right) \times D} \;\;\;\; \\
    \;\;\;\; W_{e}^{\;\text{act}} \in \mathbb{R}^{ \left( \prod_{i = 1}^N N_A(i) \right) \times D}
\end{gather}
Given the pair of morphology tokens and the pair of morphology embeddings, we use an embedding lookup operation in order to obtain the pair of observation and action token embeddings $H_{n}^{\;\text{obs}}$ and $H_{n}^{\;\text{act}}$ for each morphology. Recall from Equation~\ref{eqn:obs_act_embeddings} these have a fixed number of columns, and a variable number of rows, which depends on the dimensionality of the agent's observations and actions. 
\begin{gather}
    H_{n}^{\;\text{obs}} = \text{embedding\_lookup} \left( \; I_{n}^{\;\text{obs}}, \; W_{e}^{\;\text{obs}} \; \right) \;\;\;\; \\
    \;\;\;\; H_{n}^{\;\text{act}} = \text{embedding\_lookup} \left( \; I_{n}^{\;\text{act}}, \; W_{e}^{\;\text{act}} \; \right)
\end{gather}
These observation and action token embeddings are shown in Figure~\ref{fig:main_diagram} as a sequence of outlined rectangles on the left of the diagram. Each colored rectangle in each sequence represents a token embedding vector with cardinality $D$. Outlined yellow rectangles represent observation tokens, and outlined red rectangles represent action tokens. While this representation of morphology does not require the agent to have limbs, it requires instead a pair of morphology tokens that identify the agent's sensors and actuators. Since this representation has the advantage of differentiability, one promising option to avoid manual annotation is to infer $H_{n}^{\;\text{obs}}$ and $H_{n}^{\;\text{act}}$ using an encoder for future work.

\begin{figure*}[htbp]
    \centering
    \includegraphics[width=\linewidth]{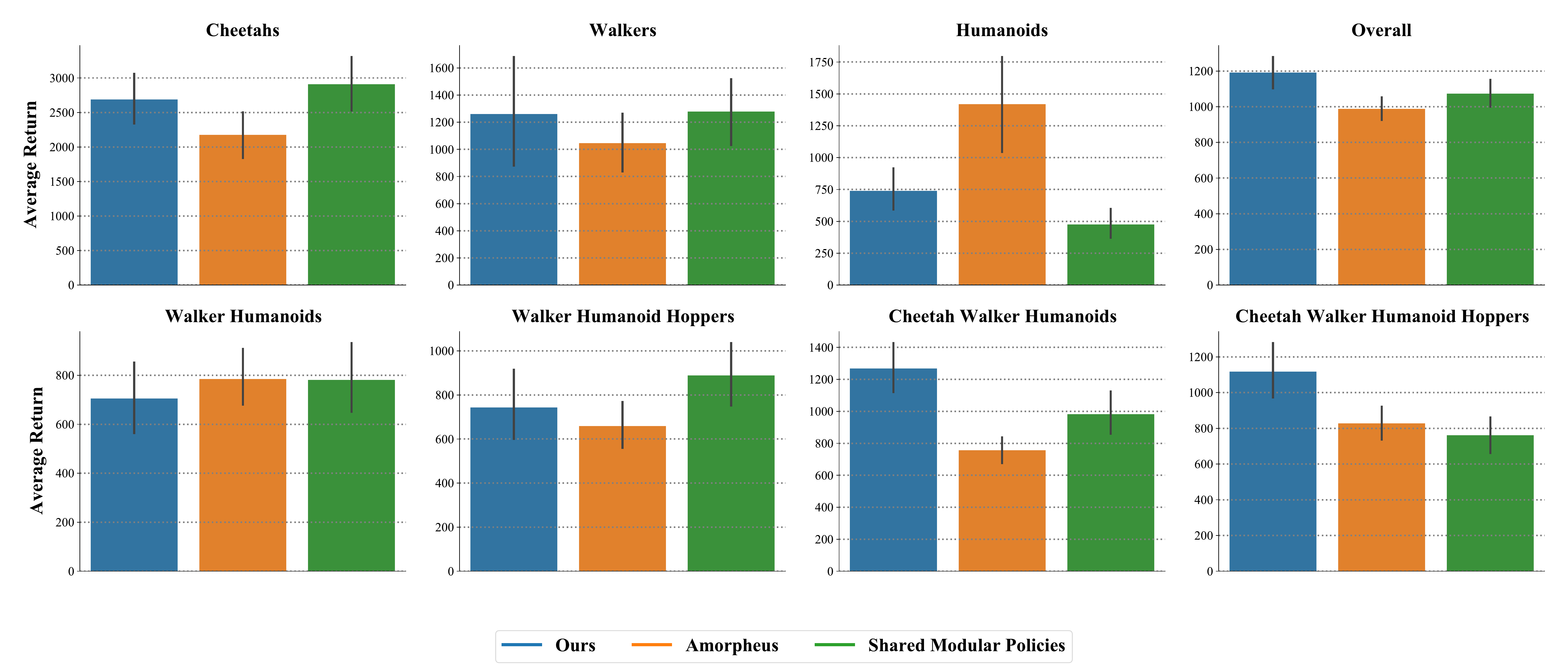}
    \vspace{-0.5cm}
    \caption{Average return of our method versus Amorpheus~\citep{DBLP:conf/iclr/KurinIRBW21} and Shared Modular Policies~\citep{DBLP:conf/icml/HuangMP20} for held-out morphologies with $4$ random seeds. Performance on the y-axis is the average return with ten episodes per held-out morphology per seed. Colored bars represent average performance evaluated at $2.5$ million environment steps, and a $95\%$ confidence interval is shown with error bars. Overall performance is an aggregation of all episodes from each method. Our model improves by $16\%$ overall, and by $28\%$ and $32\%$ on the Cheetah-Walker-Humanoids and Cheetah-Walker-Humanoid-Hoppers tasks respectively, the two hardest tasks.}
    \label{fig:section_two}
\end{figure*}

\paragraph{Embedding The Current Observation.} How can we condition our policy, which is a sequence to sequence model, on the current state? We propose to view the current state as a sequence with $N_S(n)$ elements, embedding each element to a $M$-vector, and concatenate it with the matrix of observation token embeddings $H_{n}^{\;\text{obs}}$ column-wise. We empirically find that passing the state through a sinusoidal embedding function before concatenation helps the model to perform well, which may be interpreted as increasing the expressivity of the model class \cite{li2021functional}.
Our embedding function passes the state through a series of $M$ sinusoidal functions with geometrically increasing frequencies $k_1, k_2, \hdots k_{\lfloor M / 2 \rfloor}$ from $1/10$ to $1000$. In Figure~\ref{fig:main_diagram}, we represent this operation with the yellow trapezoid labelled embed on the left, and denote the subsequent concatenation of the embedded state with $H_{n}^{\;\text{obs}}$ via the plus symbol.
\begin{equation}
    X = \Big[ \; H_{n}^{\;\text{obs}}; \; \underbrace{ \cos \left( k_1 \mathbf{s}_t \right); \; \sin \left( k_1 \mathbf{s}_t \right); \; \hdots }_{M \; \text{columns}} \; \Big]
\end{equation}
Before processing $X$ and $H_{n}^{\;\text{act}}$ with our transformer, we apply a linear transformation that maps them to the cardinality of the Transformer hidden state. We learn two linear transformations $W_{p}^{\;\text{obs}}$ and $W_{p}^{\;\text{act}}$ for observation and action embeddings respectively, projecting to $d_{\text{model}}$ components.
\begin{gather}
    W_{p}^{\;\text{obs}} \in \mathbb{R}^{(D + M) \times d_{\text{model}}}, \; W_{p}^{\;\text{act}} \in \mathbb{R}^{D \times d_{\text{model}}}
\end{gather}
After projection, we process these embeddings with a Transformer encoder-decoder model, following the architecture presented by \citet[p. 3]{DBLP:conf/nips/VaswaniSPUJGKP17}. Note that we drop autoregressive masking, which is unnecessary in our setting. Details about the hyperparameters used with our model can be found in Appendix~\ref{app:hyperparameters}. The output of our Transformer is a sequence of $N_A(n)$ hidden state vectors with $d_{\text{model}}$ components. To parameterize an action distribution, we learn a final linear transformation $W_{p}^{\;\text{out}}$ that projects from $d_{\text{model}}$-vectors to scalars. Our model outputs a vector $y_t$ with $N_A(n)$ components, representing pre-activation actions. 
\begin{gather}
    W_{p}^{\;\text{out}} \in \mathbb{R}^{d_{\text{model}}}\\
    \mathbf{y}_t = \text{Transformer} \big(\; X W_{p}^{\;\text{obs}}, \; H_{n}^{\;\text{act}} \; W_{p}^{\;\text{act}} \; \big) W_{p}^{\;\text{out}}
\end{gather}
With a final hyperbolic tangent activation, we generate the mean action $\mathbf{a}_t = \tanh( \mathbf{y}_t )$ for the current timestep. This conversion from Transformer outputs to actions is shown in Figure~\ref{fig:main_diagram} by the Linear and Squeeze boxes on the right.
By framing learning morphology as sequence modelling, our model is applicable to a larger space of agent types, given appropriate morphology tokens and sufficient training. Furthermore, our model does not assume the agent conforms to strict design criteria seen in prior work and detailed in Section~\ref{sec:morphology}. In the remaining sections, we will benchmark our model on a standard set of agent-agnostic reinforcement learning tasks, and visualize what our model learns.

\paragraph{Policy Optimization.} We choose to employ TD3 \cite{DBLP:conf/icml/FujimotoHM18} for optimizing our policy because it is highly efficient and consistently performs well for MuJoCo-like \cite{DBLP:conf/iros/TodorovET12} agents. Efficiency is an important consideration in agent-agnostic reinforcement learning because training with up to $32$ morphologies can require large amounts of experience otherwise. Following the TD3 algorithm, we sample exploratory actions from a normal distribution $\mathcal{N}(\mathbf{a}_t, \sigma^2)$ where $\mathbf{a}_t$ is the mean action, and the variance $\sigma^2$ is a hyperparameter that controls the degree of randomness in the exploration policy of TD3. Following prior work \cite{DBLP:conf/icml/HuangMP20} we store a separate replay buffer of transitions for each training morphology, and alternate between collecting environment steps and training for each morphology in lockstep. We train each agent for up to 3 million environments steps total across all training morphologies, and run $4$ random seeds per method. Our model fits on a single Nvidia 2080ti GPU, and requires seven days of training to reach 3 million environments steps. We provide a table of hyperparameters in Appendix~\ref{app:hyperparameters} for our policy and reinforcement learning optimizer. Additionally, we have released the source code for our method and summarized how the model works at the following \href{https://sites.google.com/btrabucco.com/anymorph-icml2022}{site}.

%% file: sections/experiments_one.tex
\begin{figure*}[htbp]
    \centering
    \includegraphics[width=\linewidth]{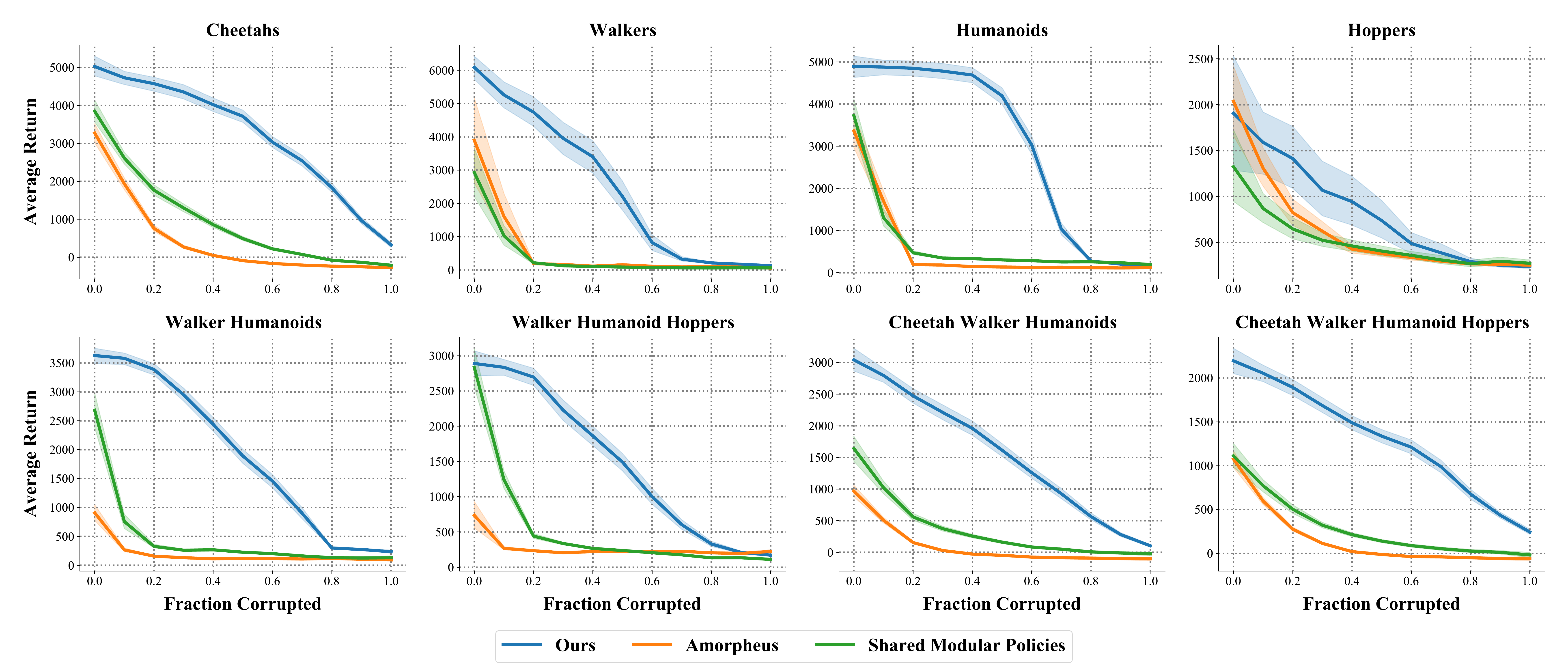}
    \caption{Average return of our method versus Amorpheus~\cite{DBLP:conf/iclr/KurinIRBW21} and Shared Modular Policies~\cite{DBLP:conf/icml/HuangMP20} as a function of how many sensors are corrupted by random noise. Performance on the y-axis is the average return of $4$ random seeds with one episode per morphology for each seed. Dark colored lines indicate the average performance for each method given a fraction of sensors that are corrupted by noise sampled from a standard normal distribution, and a $95\%$ confidence interval is shown with shading around each line. The x-axis indicates the ratio of how many sensors in the agent's observation space are corrupted with noise, divided by the total number of sensors the agent has. Our method consistently improves robustness, illustrated by a greater area under the curve.}
    \label{fig:section_three}
\end{figure*}

We have constructed a method that is agnostic the morphology of an agent. This model is trained across a collection of different agents to develop skills that generalize to new morphologies. To understand how well our method succeeds we evaluate three aspects of the model. First, how quickly can we train the morphology agnostic policy. We should receive gains in sample efficiency as the model learns how to share experience across morphologies. Second, how well does the method generalize to novel morphologies not seen during the training process. Last, how robust is our method to sensor issues. To answer these questions, we leverage a benchmark for agent-agnostic reinforcement learning developed by \citet[p. 1]{DBLP:conf/icml/HuangMP20}. This benchmark contains a set of eight reinforcement learning tasks, where the goal is to maximize the average return over a set of $N$ agents with different morphologies. The agents present in this benchmark and inspired by and derived from standard OpenAI Gym tasks: HalfCheetah-v2, Walker2d-2, Hopper-v2, and Humanoid-v2~\cite{DBLP:journals/corr/BrockmanCPSSTZ16}. For the Cheetahs, Walkers, Humanoids, and Hoppers tasks, there are a total of $15$, $6$, $8$, and $3$ different morphologies used in prior work~\cite{DBLP:conf/iclr/KurinIRBW21, DBLP:conf/icml/HuangMP20}. The benchmark consists of four tasks containing morphologies of a single \textit{kind} of agent (see Cheetahs, Walkers, Humanoids, and Hoppers in Figure~\ref{fig:section_one}), and four tasks that mix together multiple kinds of agents (see Walker-Humanoids, Walker-Humanoid-Hoppers, Cheetah-Walker-Humanoids, and Cheetah-Walker-Humanoid-Hoppers in Figure~\ref{fig:section_one}). Solving the latter mixed tasks requires the policy to acquire gaits that generalize to multiple kinds of agents.
The two hardest tasks in the benchmark are Cheetah-Walker-Humanoids, and Cheetah-Walker-Humanoid-Hoppers, which involve controlling $29$ and $32$ different agent morphologies respectively.

Our method excels at controlling many morphologies, and sees greater improvements as the amount of morphologies increases. Shown in \autoref{fig:section_one}, on all tasks but Hoppers, our policy meets and exceeds the performance of Amorpheus \cite{DBLP:conf/iclr/KurinIRBW21} and Shared Modular Policies \cite{DBLP:conf/icml/HuangMP20}, and avoids making strong assumptions about the agent's design. Furthermore, on the hardest tasks, Cheetah-Walker-Humanoids, and Cheetah-Walker-Humanoid-Hoppers, we see an improvement of $85\%$ and $53\%$
respectively. This improvement suggests that our method is more scalable than prior work, as it trains across up to $32$ different morphologies, while maintaining strong performance. Our methods performs consistently well across all tasks except Hoppers. The Hoppers task has only three morphologies, the fewest in the benchmark, and we suspect this relates to our diminished performance.

%% file: sections/experiments_two.tex
In the previous section, we evaluated the training performance of our model, and demonstrated a significant improvement when controlling many morphologies. Now we ask: how well does our model generalize to new morphologies it was not trained on? To answer this question, we follow \citet{DBLP:conf/iclr/KurinIRBW21} and hold out 3 Cheetahs, 2 Walkers, and 2 Humanoids respectively. See Appendix~\ref{app:held_out} for which specific morphologies are used for testing. We then evaluate the policies learned by our method, Amorpheus, and Shared Modular policies at $2.5$ million environment steps on each morphology that are held out. For tasks that involve multiple kinds of agents, we evaluate using all held out morphologies for each kind. We report the average return for each method over $4$ random seeds, and ten episodes per seed, with a $95\%$ confidence interval in~\autoref{fig:section_two}.

Despite not explicitly conditioning on morphology via graph structure or sensor-to-limb alignment, our method improves by $16\%$ in overall zero-shot generalization. On tasks with fewer morphologies, our methods performs on par with existing methods, with an exception on the Humanoids tasks. We obtain an improvement of $28\%$ on Cheetah-Walker-Humanoids, and $32\%$ on Cheetah-Walker-Humanoid-Hoppers, the two hardest tasks in the benchmark. The disparity between training and testing performance on certain tasks (such as the Humanoids task) suggests observing many morphologies during training is key to regularization that enables our model to generalize effectively. 

%% file: sections/experiments_three.tex
\begin{figure}
    \centering
    \includegraphics[width=\linewidth]{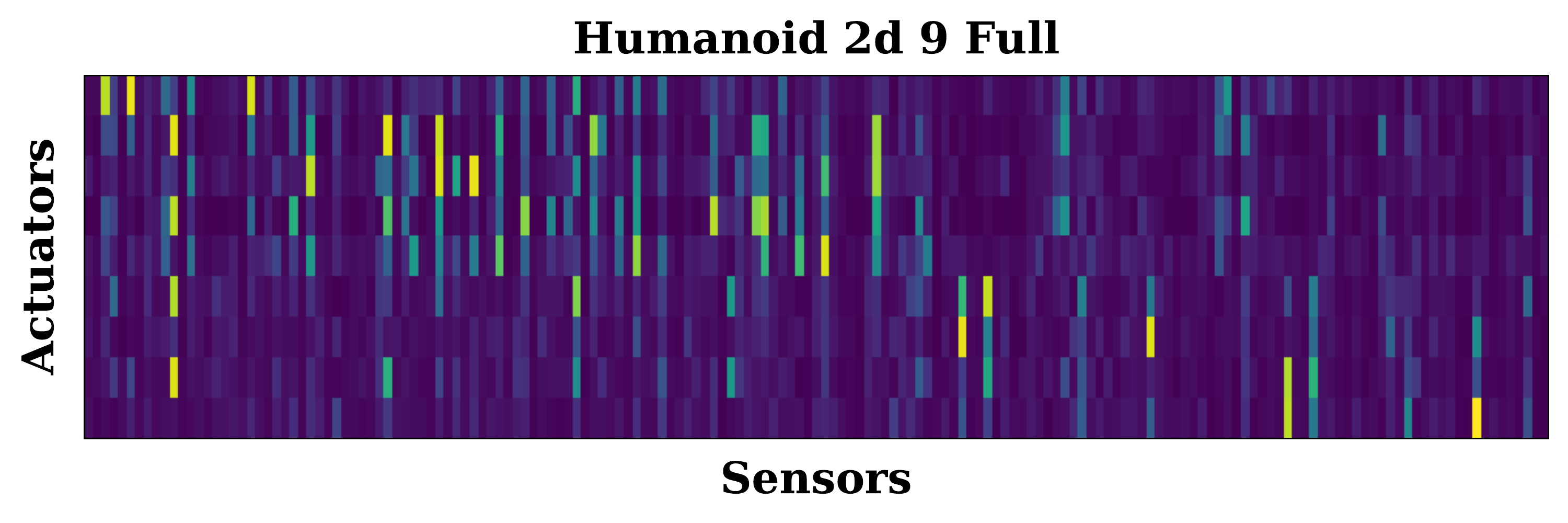}
    \vspace{-0.5cm}
    \caption{Visualization of the cross attention weights in the final layer of  our policy model on the Humanoids task and Humanoid 2d 9 Full morphology. Each cell in the attention matrix corresponds to the maximum
    attention value for that cell over an entire episode. Cells with dark shading indicate that our model ignores those sensors over an entire episode by zeroing their attention weights.}
    \label{fig:attention_masks}
\end{figure}

In the two previous experiments, we evaluated the training performance and zero-shot generalization ability of our model, and displayed compelling gains. One remaining question, however, is how resilient our model is to \textit{minor} changes in the agent's morphology, caused by one or more sensors breaking. We investigate this question by designing an experiment where a fraction of the agent's sensors are replaced with noise sampled from the standard normal distribution. The goal of this experiment is to evaluate the robustness of our model, measured by average return, as a function of how many of the agent's sensor readings (state features) are replaced with random noise. Our methodology for selecting the order to corrupt sensors is described in Appendix~\ref{app:robustness}. An evaluation of the robustness of our method compared to prior works is presented in~\autoref{fig:section_three}, where the average return for each method given a certain fraction of corrupted sensors is calculated from one episode per morphology per task using deterministic actions, with 4 random seeds per method on each task. Training average return is plotted on the y-axis, with a $95\%$ confidence interval.

This experiment demonstrates that our policy architecture is more robust to broken sensors than prior works. Our policy achieves a higher area under the curve on all eight tasks. Though our method initially performs worse than Amorpheus (illustrated in~\autoref{fig:section_one}) on the Hoppers task, broken sensors quickly cripple existing methods. One promising hypothesis for this improved robustness is that our model is actively ignoring certain sensors. To evaluate this hypothesis, we visualize the attention weights throughout an episode in the final cross attention mechanism in our model (that is, actuator-to-sensor attention) in~\autoref{fig:attention_masks}. For each cell in the visualization, we take the maximum attention weight over an entire episode (rows no longer sum to one) for one trial on the Humanoids task and Humanoid 2d Full morphology. This visualization shows that our model learns sparse attention weights that ignore the majority of sensors, shown by the majority of dark cells in the visualization. This quality in the attention weights suggests our model improves robustness by sparsely attending to sensors.

%% file: sections/conclusion.tex
We have presented a method for learning transferable policies between agents of different morphology, by inferring the agent's morphology via a learned embedding. Our approach is more scalable than existing methods, and is able to learn composable polices for up to $32$ different morphologies at once while maintaining a performance lead of $82\%$ in training, and $32\%$ in zero-shot generalization on our benchmarking task with the most morphologies. Our method operates by framing learning morphology as a sequence modelling problem, and learns a Transformer-based policy that is invariant to the dimensionality of the agent's observations and actions. In addition to improving performance, we demonstrate that our policy is more resilient to broken sensors than existing methods. Importantly, our method attains these improvements while also relaxing the amount of information required about the agent's design compared to prior work. Our method does not require the agent to have limbs, graph structure, or aligned sensors and actuators. By relaxing these assumptions, our approach helps to improve the applicability of agent-agnostic learning in a more general reinforcement learning context.

Our research is a step towards broadly applicable agent-agnostic reinforcement learning methods, and there are several opportunities to expand our work. Firstly, we observed in Section~\ref{sec:experiments_two} that our methods benefits from observing many morphologies during training. Further scaling of our method to agent-agnostic reinforcement learning tasks with significantly more morphologies may further improve zero-shot generalization. Secondly, we evaluated our method on MuJoCo-like agents with rigid limbs in this work. Applying our method to reinforcement learning tasks \textit{without} an underlying graph structure or aligned sensors and limbs (such as Atari) poses an interesting challenge. Finally, our method requires morphology tokens (see Section~\ref{sec:morphology}) for each agent. Inferring these tokens from trajectories the agent has collected may enable our method to generalize out-of-the-box without requiring any manual annotation of the task.

%% file: sections/appendix.tex
\section{Hyperparameters}
\label{app:hyperparameters}

In this section, we describe the hyperparameters used with our model. These include hyperparameters in our model architecture, as well as hyperparameters for the TD3 optimizer that we use to optimize our policy. The hyperparameters for our model architecture can be found in the below table, while hyperparameters for TD3 can be found two tables below.

\begin{table}[h]
    \centering
    \caption{Hyperparameters for our model architecture.}
    \vspace{0.5cm}
    \begin{tabular}{l|r}
        \toprule
        \textbf{Hyperparameter Name} & \textbf{Hyperparameter Value} \\
        \midrule
        $D$ (Token Embedding Size) & 32 \\
        $M$ (Sinusoidal Embedding Size) & 96 \\
        Transformer Hidden Size & 128 \\
        Transformer Feedforward Size & 256 \\
        Attention Heads & 2 \\
        Transformer Encoder Layers & 3 \\
        Transformer Decoder Layers & 3 \\
        Transformer Activation & relu \\
        Dropout Rate & 0.0 \\
        \bottomrule
    \end{tabular} 
    \label{tab:hyperparameters_model}
\end{table}

In the below table, we report the standard hyperparameters that are typically exposed to the user in TD3. Note that we employ the same TD3 hyperparameters across every task, which demonstrates our model does not require per-task tuning.

\begin{table}[h]
    \centering
    \caption{Hyperparameters for our TD3 implementation.}
    \vspace{0.5cm}
    \begin{tabular}{l|r}
        \toprule
        \textbf{Hyperparameter Name} & \textbf{Hyperparameter Value} \\
        \midrule
        Num Random Seeds & 4 \\
        Batch Size & 100 \\
        Max Episode Length & 1000 \\
        Max Replay Size Total & 10000000 \\
        Max Environment Steps & 3000000 \\
        Policy Update Interval & 2 \\
        Initial Exploration Steps & 10000 \\
        Policy Noise & 0.2 \\
        Policy Noise Clip & 0.5 \\
        $\tau$ & 0.046 \\
        $\sigma$ & 0.126 \\
        Discount Factor & 0.99 \\
        Gradient Clipping & 0.1 \\
        Learning Rate & 0.00005 \\
        \bottomrule
    \end{tabular}
    \label{tab:hyperparameters_td3}
\end{table}

We use identical hyperparameters for the Amorpheus \cite{DBLP:conf/iclr/KurinIRBW21} and Shared Modular Policies \cite{DBLP:conf/icml/HuangMP20} baselines, except we update the learning rate to $0.0001$, which we found to result in the best performance for both methods. We use the same reinforcement learning framework as \citet{DBLP:conf/iclr/KurinIRBW21} and  \citet{DBLP:conf/icml/HuangMP20} for our experiments.

\section{Details Of Robustness Experiment}
\label{app:robustness}

In the main paper, we performed an experiment testing the robustness of our approach versus Amorpheus \cite{DBLP:conf/iclr/KurinIRBW21} and Shared Modular Policies \cite{DBLP:conf/icml/HuangMP20} baselines. In order to determine the order in which sensors "break," simulated by replacing sensor readings with random noise sampled from the standard normal distribution, we visualized the cross attention weights in the final decoder layer of our Transformer policy, and sorted the sensors in increasing order according to how frequently our policy attends to them throughout an episode. Specifically, we take the average value of the final cross attention mask throughout an episode, and average again over the queries axis, in order to obtain a vector with $N_S(n)$ elements, representing the average attention weight applied to a given sensor by \textit{any} actuator through an episode. In order to corrupt a particular fraction $c$ of the agent's sensors, we replace the first $\lceil c \cdot N_S(n) \rceil$ sensors with random noise, according to their sorted order as previously described. This methodology ensures that sensor corruptions are cumulative. That is, for two given fractions $c_1 > c_0$, the sensors corrupted by $c_0$ are also corrupted by $c_1$.

\section{Held Out Morphologies}
\label{app:held_out}

In this section, we present a table that shows which morphologies are used for training and which morphologies are used for testing for each kind of agent. For tasks that mix multiple kinds of agents, the training morphologies for each kind are mixed, and the testing morphologies are mixed, but no training morphology becomes a testing morphology and vice versa.

\begin{table}[h]
    \centering
    \caption{Morphologies used for training and testing.}
    \vspace{0.5cm}
    \begin{tabular}{l|l|l}
        \toprule
        \textbf{Task} & \textbf{Training Morphologies} & \textbf{Testing Morphologies} \\
        \midrule
        Cheetahs & & \\
        & cheetah\_2\_back & cheetah\_3\_balanced \\
        & cheetah\_2\_front & cheetah\_5\_back \\
        & cheetah\_3\_back & cheetah\_6\_front \\
        & cheetah\_3\_front & \\
        & cheetah\_4\_allback & \\
        & cheetah\_4\_allfront & \\
        & cheetah\_4\_back & \\
        & cheetah\_4\_front & \\
        & cheetah\_5\_balanced & \\
        & cheetah\_5\_front & \\
        & cheetah\_6\_back & \\
        & cheetah\_7\_full & \\
        \midrule
        Walkers & & \\
        & walker\_2\_main & walker\_3\_main \\
        & walker\_4\_main & walker\_6\_main \\
        & walker\_5\_main & \\
        & walker\_7\_main & \\
        \midrule
        Humanoids & & \\
        & humanoid\_2d\_7\_left\_arm & humanoid\_2d\_7\_left\_leg \\
        & humanoid\_2d\_7\_lower\_arms & humanoid\_2d\_8\_right\_knee \\
        & humanoid\_2d\_7\_right arm & \\
        & humanoid\_2d\_7\_right leg & \\
        & humanoid\_2d\_8\_left knee & \\
        & humanoid\_2d\_9\_full & \\
        \midrule
        Hoppers & & \\
        & hopper\_3 & \\
        & hopper\_4 & \\
        & hopper\_5 & \\
        \bottomrule
    \end{tabular}
    \label{tab:held_out_morphologies}
\end{table}

\section{Morphology Token Selection}
\label{app:tokens}

We document the integer values selected for the morphology tokens used in our paper for every full morphology in the benchmark provided by \cite{DBLP:conf/icml/HuangMP20}. These full morphologies are \texttt{cheetah\_7\_full},  \texttt{walker\_7\_main},  \texttt{humanoid\_2d\_9\_full} and \texttt{hopper\_5}. Documenting morphology tokens for these is sufficient to reproduce the paper because, for any sub-morphology that has one or more limbs missing, such as \texttt{cheetah\_6\_back} which is missing the limb titled \textit{ffoot}, the morphology tokens corresponding to sensors and motors attatched to the missing limb(s) are withheld.

\begin{table}[h]
    \centering
    \caption{Morphology tokens for the observation spaces of every task. The variable $i$ indicates that each of the agent's limbs has all of the above sensors, and each limb's sensor has a unique consequently increasing integer token value assigned to it. The range of token values provided illustrates that certain sensors produce vector observations, such as the position sensor, which produces three values.}
    \vspace{0.5cm}
    \begin{tabular}{l|r}
        \toprule
        \textbf{Sensor Name} & \textbf{Token Integer Range} \\
        \midrule
        Limb[$i$] Position & $[18i+0, 18i+3)$ \\
        Limb[$i$] Orientation & $[18i+3, 18i+6)$ \\
        Limb[$i$] dPosition / dt & $[18i+6, 18i+9)$ \\
        Limb[$i$] dOrientation / dt & $[18i+9, 18i+12)$ \\
        Limb[$i$] Type & $[18i+12, 18i+16)$ \\
        Limb[$i$] Angle & $[18i+16, 18i+17)$ \\
        Limb[$i$] Joint Range & $[18i+17, 18i+18)$ \\
        \bottomrule
    \end{tabular}
    \label{tab:observation_tokens}
\end{table}

Table~\ref{tab:observation_tokens} shows how observation tokens are selected for a single kind of morphology. In particular, the table shows how tokens are assigned for the \textit{full} morphology in each family. For each sub-morphology that has one or more limbs missing, the sensor observations connected to that limb are missing from the agent's total observation, and token values for that limb are similarly withheld. When selecting observation token values for sensors with multiple kinds of morphologies present, such as Humanoids and Walkers, we follow an identical process with an additional step of adding a predetermined constant to the observation token values of each morphology kind to ensure no collisions occur. For example, suppose the  \texttt{humanoid\_2d\_9\_full} morphology has $m_0$ consecutive observation tokens values assigned to it, starting from zero. We then add $m_0$ to every observation token value for the \texttt{walker\_7\_main} morphology. If the \texttt{walker\_7\_main} morphology has $m_1$ consecutive observation tokens values assigned to it, then combining these two kinds of morphologies results in $m_0 + m_1$ unique token values. This ensures no sensors share the same token value between kinds of morphologies.

\begin{table}[h]
    \centering
    \caption{Morphology tokens for the action spaces of every task.}
    \vspace{0.5cm}
    \begin{tabular}{l|r||l|r}
        \toprule
        \textbf{Motor Name} & \textbf{Token Integer Value} & \textbf{Motor Name} & \textbf{Token Integer Value} \\
        \midrule
        Cheetah FFoot & $0$ & Walker Left1 & $0$ \\
        Cheetah FShin & $1$ & Walker Left2 & $1$ \\
        Cheetah FThigh & $2$ & Walker Left3 & $2$ \\
        Cheetah BFoot & $3$ & Walker Right1 & $3$ \\
        Cheetah BShin & $4$ & Walker Right2 & $4$ \\
        Cheetah BThigh & $5$ & Walker Right3 & $5$ \\
        \bottomrule
        \toprule
        \textbf{Motor Name} & \textbf{Token Integer Value} & \textbf{Motor Name} & \textbf{Token Integer Value} \\
        \midrule
        Humanoid Left Shoulder & $0$ & Hopper Thigh & $0$ \\
        Humanoid Left Elbow & $1$ & Hopper Leg & $1$ \\
        Humanoid Left Hip & $2$ & Hopper Lower Leg & $2$ \\
        Humanoid Left Knee & $3$ & Hopper Foot & $3$ \\
        Humanoid Right Shoulder & $4$ & & \\
        Humanoid Right Elbow & $5$ & & \\
        Humanoid Right Hip & $6$ & & \\
        Humanoid Right Knee & $7$ & & \\
        \bottomrule
    \end{tabular}
    \label{tab:action_tokens}
\end{table}

Table~\ref{tab:action_tokens} shows how action tokens are selected for a single kind of morphology. We follow a similar procedure for selecting action tokens and observations tokens, which assigns consecutive integers token values to each motor in the \textit{full} morphology for each family. For each sub-morphology that has one or more limbs missing, the motor actions connected to that limb are missing from the agent's total action, and action token values for that limb are similarly withheld. When selecting action token values for motors with multiple kinds of morphologies present, such as Humanoids and Walkers, we follow the previously described procedure for selecting observation tokens, where we add a predetermined constant (a different constant than for observation tokens) to the action token values of each morphology kind to ensure no collisions occur. 

Our results show this procedure is effective for MuJoCo-like \citep{DBLP:conf/iros/TodorovET12} agents, but this procedure, in its simplicity, has several limitations. First, since no token collisions occur between kinds of morphologies, the token selection method described above does not support \textit{out-of-domain} zero-shot generalization without modification. This is the kind of zero-shot generalization necessary when training our method on the Humanoid family of morphologies, and testing it on the Walker family of morphologies, without training our method on a single walker morpohology. Notably, prior works in agent-agnostic RL, including \cite{DBLP:conf/icml/HuangMP20} and \cite{DBLP:conf/iclr/KurinIRBW21} only consider \textit{in-domain} zero-shot generalization, such as training on non-empty subsets of both Humanoid morphologies and Walker morphologies, and testing on different subsets, allowing the policy to experience at least one example each of a humanoid and walker during training.

Evaluating our method on \textit{out-of-domain} zero-shot generalization requires modifying how observation and action tokens are selected, to ensure the learned morphology representation can be transferred to novel morphology families. Effective token selection methods for \textit{out-of-domain} zero-shot generalization likely need some degree of token collision to occur between morphology families. Further research into how to select the morphology tokens to enable \textit{out-of-domain} zero-shot generalization is a promising research direction, and is a necessary step to apply our method to more realistic RL problems.